\pdfoutput=1

\documentclass[11pt]{article}

\usepackage[]{ACL2023}

\usepackage{times}
\usepackage{latexsym}

\usepackage[T1]{fontenc}

\usepackage[utf8]{inputenc}

\usepackage{microtype}

\usepackage{inconsolata}

\usepackage{float}
\usepackage{multirow}
\usepackage{tikz, pgf, pgfplots, pgfplotstable}
\usepackage{amsmath}
\usepackage{booktabs}
\usepackage{comment}

\newcommand\checkthis[1]{\textcolor{black}{#1}}
\newcommand\scalemath[2]{\scalebox{#1}{\mbox{\ensuremath{\displaystyle #2}}}}

\title{
Contextual Knowledge Learning For Dialogue Generation} 

\author{Wen Zheng \\
  University of Nottingham  \\
  \text{wen.zheng@nottingham.ac.uk} \\
  \And
  Natasa Milic-Frayling \\
  University of Nottingham \\
  \text{natasa.milic-frayling@nottingham.ac.uk} \\
  \AND 
  Ke Zhou \\
  University of Nottingham \& Nokia Bell Labs \\
  \text{ke.zhou@nottingham.ac.uk} \\
}

\begin{document}
\maketitle
\begin{abstract}

Incorporating conversational context and knowledge into dialogue generation models has been essential for improving the quality of the generated responses. The context, comprising utterances from previous dialogue exchanges, is used as a source of content for response generation and as a means of selecting external knowledge. However, to avoid introducing irrelevant content, it is key to enable fine-grained scoring of context and knowledge. In this paper, we present a novel approach to context and knowledge weighting as an integral part of model training.
We guide the model training through a Contextual Knowledge Learning (CKL) process which involves Latent Vectors for context and knowledge, respectively. CKL Latent Vectors capture the relationship between context, knowledge, and responses through weak supervision and enable differential weighting of  context utterances and knowledge sentences during the training process. Experiments with two standard datasets and human evaluation demonstrate that CKL leads to a significant improvement compared with the performance of six strong baseline models and shows robustness with regard to reduced sizes of training sets.

\end{abstract}

\section{Introduction}

Dialogue generation is concerned with conversational settings where participants take turns and the task is to generate a response to previous utterances. In order to generate relevant responses, prior research explored the use of the conversational history, i.e., the utterances already exchanged between the participants, as a context for the new response. It also investigated the use of external knowledge that may be a good source of content and not necessarily present in the conversational history \citep{firstknowledge_Ghazvininejad, ted_zheng2019enhancing, liu2021pre, kat-liu2021three, doha-prabhumoye2021focused}.
However, it has been shown that adding knowledge indiscriminately, can hurt performance. Thus, the context has been used to select the best knowledge for the response generation.

\begin{figure}[t]
  \centering
  \includegraphics[width=0.95\linewidth]{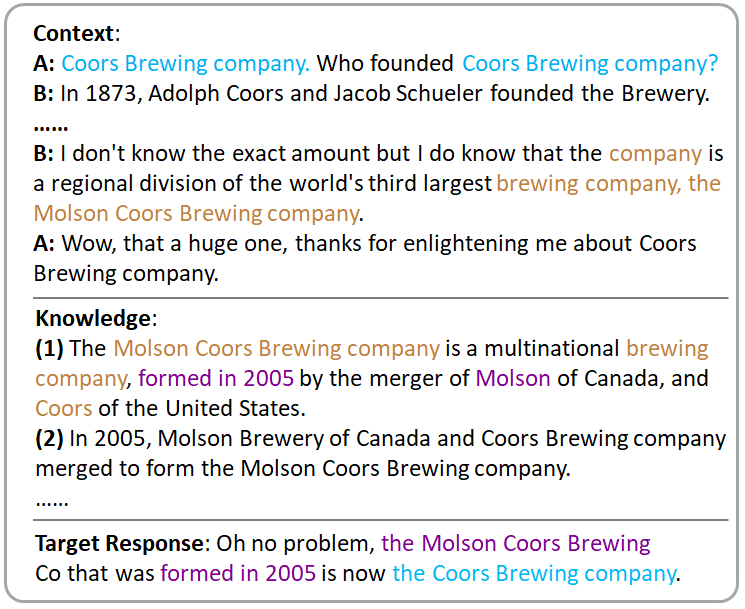}
  \caption{Example showing utterances of participants A and B, the scored knowledge, and the target response. The knowledge sentence (1) is deemed the best knowledge for response generation by our proposed CKL model. The best context segments for retrieving the best knowledge are colored \textcolor{brown}{Brown}; the best context segments for generating response are colored \textcolor{cyan}{Blue} and the best knowledge segments for the response are \textcolor{violet}{Purple}.}
  \label{fig:example}
\vspace{-0.1in}
\end{figure} 

Since the context itself consists of multiple utterances, the same concern applies: not all the prior utterances are equally useful for generating the response. Therefore, the context needs to be evaluated for its importance in relation to generating the response and identifying the relevant knowledge, separately. 
In Figure~\ref{fig:example}, we show an example from the Wizard of Wikipedia (test seen) dataset, illustrating that the best context segments for response generation may not necessarily be the best context segments for retrieving the best knowledge.
Furthermore, both the context and the contextual knowledge contribute to the coverage of the target response (\textcolor{cyan}{blue} and \textcolor{violet}{purple} words). Thus, it is important to devise effective learning methods to identify the best context for response generation and for knowledge selection. Once the knowledge is selected, there is still a question of whether and how to refine its selection for optimal use.

{Recent studies differentiated between the two context roles by adopting a pipeline approach and training different models for each of them.} 
{\citet{tppa_zheng2020approximation} proposed a knowledge retrieval model TPPA that re-orders retrieved knowledge guided by its relevance to the response and investigated the effects of the resulting knowledge sets in combination with generative models such as TED \citep{ted_zheng2019enhancing} and WSeq \citep{wseq_tian2017make}. \citet{paranjape2021hindsight} introduced a posterior-guided training to guide the retrieval of the relevant knowledge. A BART-based generative model (a generator) is used to generate responses but the retriever and the generator are trained independently. Similarly, \citet{glass2022re2g} developed a $Re^2G$ model which comprises a retriever, a re-ranker, and a generator. The re-ranker can take as input the outputs of multiple retrieval systems, e.g., ANN-based retrieval and BM25 method and the content retrieval training is an integral part of the content generation. This approach can differentiate the context roles from knowledge selection and response generation tasks. However, it requires additional training stages, which may incur and accumulate additional errors, and cannot separate the context information used for knowledge selection and response generation within the unified model.}

In our work we hypothesize that the integrated approach to model training and selection of context and knowledge can be improved through a parallel learning architecture where specific content selection roles (context and knowledge) are clearly differentiated and each learning facet is supervised, controlling for model training. Guided by the hypothesis, we propose a Contextual Knowledge Learning ({\bf CKL}) model in which we introduce \textit{Latent Vectors} to capture context roles and knowledge characteristics: the \emph{Context Latent Vector} for the relationship of context to the responses and to the `best' knowledge, and the \emph{Knowledge Latent Vector} for the knowledge to capture the importance of knowledge to the responses. 
\textit{Latent Weights} are then derived from the Latent Vectors to indicate the importance of context utterances and knowledge sentences.
%
%

{We also extend the notion of the Attention operation, where tokens' attention scores are entirely decided by the scaled dot product between two representations, and devise a \textit{Latent Weight Enhanced Attention}. The attention operation is augmented with the multiplication by the tokens' attention scores and the \emph{Latent Weights} (i.e., the context utterance's weight and knowledge sentence weight). By adopting the weak supervision technique, the \emph{Latent Weights} for context and knowledge are supervised by the (noisy) pseudo ground truth, removing the need for human annotations. Combined with the Negative Log Likelihood loss, the CKL is trained in a unified way, differentiating the context utterances for the knowledge selection and response generation tasks. } 

The performance that our CKL model is superior to six strong baseline approaches, including Transformer-based and pre-trained model-based methods, on two publicly available datasets Wizard of Wikipedia and CM-DoG. By experimenting with a 50\% smaller training set, our approach still outperforms the baseline methods. 
Figure~\ref{fig:example} shows the effectiveness of recognizing relevant context utterances for both the knowledge selection task and dialogue generation task using the CKL model.

In summary, the key contributions of our research are through the novel design and clear advantages of the CKL method\footnote{\url{https://github.com/tonywenuon/acl2023-ckl}}: 
\begin{itemize}
{\item Differentiated functionality of the context utterances for the knowledge selection task and response generation task, achieved through the technique of training latent vector;
\item  Latent Weight Enhanced Attention module that incorporates the latent weights into the generation process;
\item Effective weak supervision of latent weights training by defining the pseudo ground truths for the context latent weights and knowledge latent weights;
\item Robustness of CKL, retaining its effectiveness with reduced amounts of data.}

\end{itemize}


\section{Related Work}


\noindent {\bf Knowledge-Grounded Dialogue Generation.}
Research on knowledge injection into dialogue generation can be traced to \citet{firstknowledge_Ghazvininejad} who demonstrated that injecting knowledge into the generative model benefits the performance due to additional information available in knowledge sentences. This led to a range of methods for knowledge injection \citep{ted_zheng2019enhancing, zhao2020knowledge, itdd_li2019incremental}. Recently, pre-trained models were also adopted. \citet{kat-liu2021three} use BART \cite{bart-lewis2020bart} as the backbone to fine-tune the model with few-resource datasets. \citet{doha-prabhumoye2021focused} also chose BART as the basic pre-trained framework to project context and knowledge in a unified model. 
{However, it has been shown that not all the knowledge is useful, and therefore knowledge has to be carefully selected \cite{skt-Kim-2020-ICLR, postks-lian2019learning}. \citet{lewis2020retrieval} propose a RAG model by leveraging retrieval techniques to obtain relevant knowledge for enhancing dialogue generation. Built on top of the RAG, \citet{shuster2021retrieval} study various types of architectures with multiple components, including retrievers, rankers, and encoder-decoders. All previously mentioned approaches take the same context information for both the knowledge selection task and the dialogue generation task. To differentiate the functionality of the context for the two tasks, \citet{glass2022re2g, paranjape2021hindsight, tppa_zheng2020approximation} introduce pipelines to achieve multi-stage generative models. Commonly, they devise a retriever that takes responsibility of retrieving knowledge units given the context information and follow by a generator that generates the final responses. The limitation is that this multi-stage training may cause error accumulation. Motivated by this, we train our CKL model in an end-to-end way by differentiating the context inside the model.}

\noindent {\bf Weak Supervision for Dialogue Generation.}
{For the dialogue generation task, even though the responses are naturally the ground truth for the model training, previous studies showed that several auxiliary tasks with weak supervision can help to improve the generation performance \citep{chang2021jointly}. \citet{zhao2020knowledge} uses BERT encoders \cite{bert-kenton2019bert} and a GPT-2 decoder \cite{gpt2-radford2019language} for knowledge projection and prediction. They built pseudo ground truth documents by leveraging the similarity between each document and the response to weakly supervise document selection. 
\citet{ktwm-zheng2021knowledge} propose a de-noising mechanism for the knowledge tokens to be injected in a weak supervision manner. \citet{wang2019weak} design a weak supervision-based discriminator to capture the relations between the answer and the corresponding passage and eventually generate the questions. Informed by these studies, in our work we introduce a weak supervision to complement the generation model training.}

{Our proposed CKL model combines context differentiation and weak supervision for response generation purposes. The CKL model differs from prior studies by (1) devising latent vectors for both context and knowledge to derive the context latent weights and knowledge latent weights for knowledge selection and response generation tasks, and (2) designing a latent weight enhanced attention operation, combined with a weak supervision technique to provide more effective use of context and knowledge for response generation.}


\begin{figure*}
  \centering
  \includegraphics[width=0.98\textwidth]{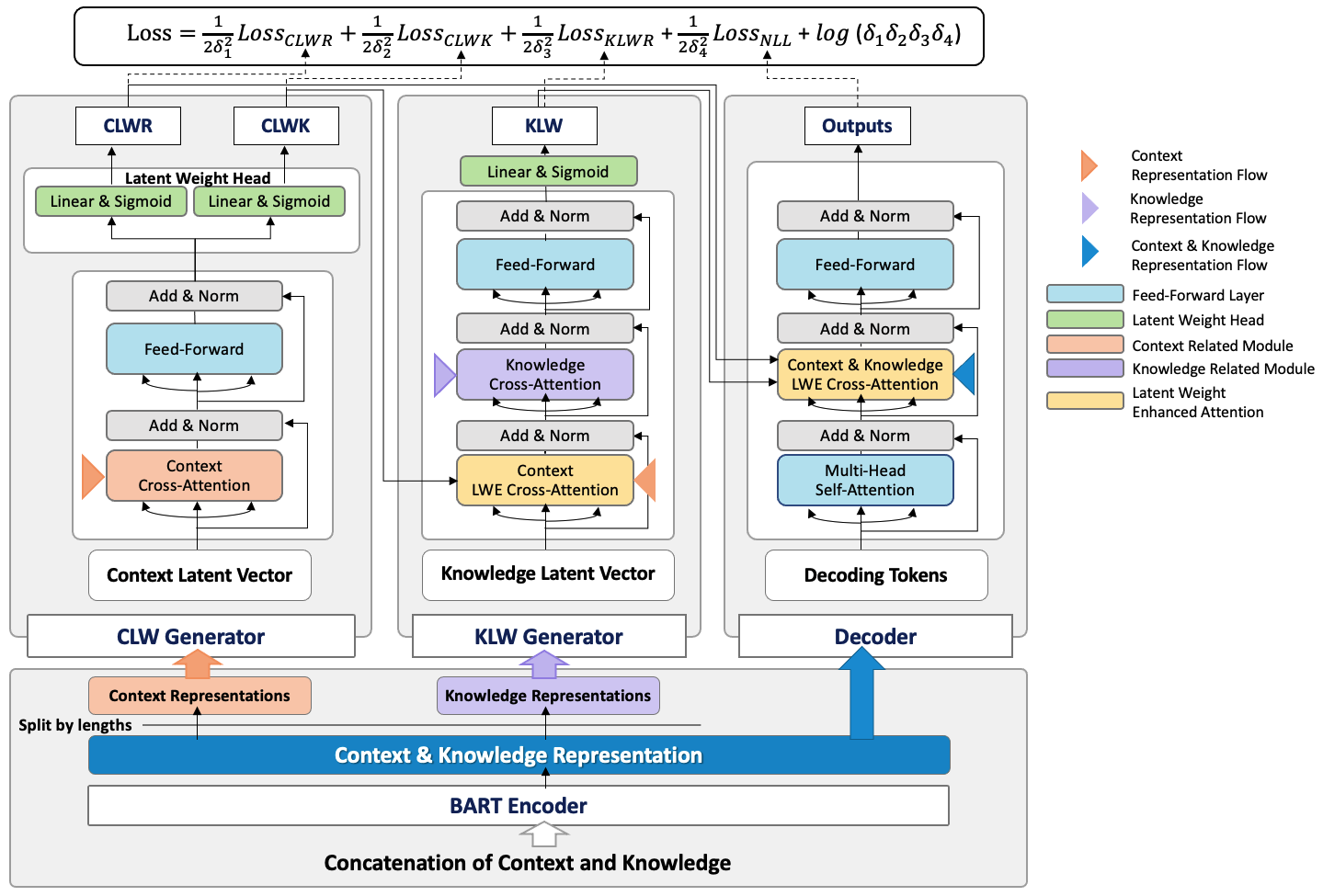}
  \caption{Contextual Knowledge Learning (CKL) model architecture comprising the BART-based encoder, two generators for latent vectors training and latent weights generation, and decoder with context and knowledge scoring. Trainable parameters $\delta_i$ balance multi-losses.}
  \label{fig:architecture}
  \vspace{-0.1in}
\end{figure*}

\section{Method} \label{sec:method}

\noindent \textbf{Problem and Definitions.} Considering a conversational history that comprises context $C = \{c_1, c_2, \dots, c_m\}$, our goal is to generate a response $R = \{r_1, r_2, \dots , r_L\}$ by leveraging knowledge $K = \{k_1, k_2, \dots , k_l\}$ that is relevant to the context $C$. Among the notations, $r_i$ is each word of the response, $c_i$ means the context utterance, and $k_i$ denotes the knowledge sentence. $L$ is the maximum token number of the response; $m$ is the number of context utterances and $l$ is the number of background knowledge sentences.

{We aim to (1) calculate latent weights of context utterances and knowledge sentences (Sec.~\ref{sec:cwpg} and~\ref{sec:kwp_generator}); (2) generate the final response given context, knowledge, and their latent weights (Sec.~\ref{sec:decoder}). In our approach, we do not integrate latent vectors into the content representation. Instead, we transform vectors into scalar values, referred to as \textit{latent weights}. The Context Latent Weights for Response and Knowledge, (\textit{CLWR}) and (\textit{CLWK}), respectively, are used in the loss function and by the decoder to score content utterances. The Contextual Knowledge Latent Weights (\textit{KLW}) are similarly used in the loss function and the decoder to score knowledge sentences.  }

\vspace{0.05in}
\noindent {\bf Contextual Knowledge Learning (CKL) Architecture.}
Our proposed CKL method consists of four components: an encoder, a Context Latent Weight generator (\textit{CLW} Generator in Figure~\ref{fig:architecture}), a Knowledge Latent Weight generator (\textit{KLW} Generator), and a decoder. We use the state-of-the-art Transformer-based encoder-decoder model BART \cite{bart-lewis2020bart} as the backbone of our Encoder and Decoder. The \textit{CLW} generator takes responsibility for producing two sets of context latent weights, one set for response generation (\textit{CLWR}) and another set for knowledge latent weight generation (\textit{CLWK}). Similar to the \textit{CLW} generator, the \textit{KLW} generator is used to generate knowledge latent weights (\textit{KLW}) which are conditioned on the context and knowledge. Finally, the decoder is a normal BART decoder but equipped with the latent weight enhanced attention mechanism. 

\subsection{Encoder}
Leveraging the pre-trained model BART, we directly use the BART encoder to get the context and knowledge representations.
The proposed CKL model needs context utterances' and knowledge sentences' representations, so they are expected to be passed through the BART encoder sequence by sequence. However, that would destroy the inner dependency between words from sequences, i.e., this means discarding the long dependency between context and knowledge. To tackle this, we first inject the concatenation of the context and knowledge to get a whole sequence representation, i.e., `Context \& Knowledge Representation' in Figure~\ref{fig:architecture}. Then by recognizing the context utterances' lengths and knowledge sentences' lengths, we split the whole representation into several sub-sequences, obtaining representations that take word long-dependency into account.

\subsection{Context Latent Weight Generator} \label{sec:cwpg}
As shown in Figure~\ref{fig:architecture}, \textit{CLW} Generator is designed to generate two sets of context latent weights: Context Latent Weight for Response (\textit{CLWR}) and Context Latent Weight for Knowledge (\textit{CLWK}).

\vspace{0.05in}
\noindent {\bf Context Latent Vector.}
The \textit{CLW} generator starts from a Context Latent Vector which is a trainable vector. Practically, it is a word embedding indexed by a fixed word index of 1. 

\vspace{0.05in}
\noindent{\bf Context Latent Vector Interaction with Context Representations.}
Here in the \textit{CLW} generator, like the Transformer architecture, a standard cross-attention, feed-forward, and residual network are used. We introduce the cross-attention operation in detail because it will be used in the next sections. For the rest of the Transformer modules, e.g., feed-forward layer and residual network, please refer to \citet{transformer-vaswani2017attention}. Formally, the attention is calculated as:
\begin{equation} \label{eq:attention}
\small
    \scalemath{1.0}{\textit{Attention(Q, K, V) = softmax} \left ( \frac{QK^T}{\sqrt{d}} \right ) V}
\end{equation}
in which Q, K, and V are matrices and $d$ is the representation dimension. Through the softmax function, the attention weights, i.e., $\textit{QK}^\textit{T}$, are normalized. Multiplying with V, the Q's representation is updated by K and V. In the \textit{CLW} generator, Q is the context latent vector, while K and V are the context representation.

\vspace{0.05in}
\noindent{\bf Latent Weight Head.}
The Latent Weight Head module contains a linear layer and a Sigmoid function. The purpose is to transfer the $d$ dimensional context latent vector to scalar values. By doing so, each context utterance will have a latent score. To be specific, we define the Latent Weight Head as follows:
\begin{equation} \label{eq:wp_head}
\small
    \textit{CLWR = Sigmoid}(x_{\textit{CLV}}W_1 + b_1)
\end{equation}
\begin{equation} \label{eq:wp_head2}
\small
    \scalemath{1.0}{\textit{CLWK = Sigmoid}(x_{\textit{CLV}}W_2 + b_2)}
\end{equation}
where $\textit{CLWR} \in \mathcal{R}^{1 \times m}$ and $\textit{CLWK} \in \mathcal{R}^{1 \times m}$ are the context latent weight scores for the response and knowledge respectively. $\textit{W}_1, \textit{W}_2 \in \mathcal{R}^{d \times 1}$ and $b_1, b_2 \in \mathcal{R}^1$ are trainable parameters. $x_{\textit{CLV}} \in \mathcal{R}^{1 \times d}$ denotes the vector converted from the Context Latent Vector. It is important to note that \textit{CLWR} and \textit{CLWK} have the same Latent Weight Head architecture, but do not share parameters. \textit{CLWR} is used to identify the importance of a context utterance when generating responses and \textit{CLWK} is used when producing knowledge latent weights, i.e., knowledge sentences' importance.

\vspace{0.05in}
\noindent {\bf Weak Supervision on \textit{CLWR} and \textit{CLWK}.}
As illustrated, when predicting the response and producing the knowledge latent weight, the role of the context utterances should not be treated as the same: \textit{CLWR} and \textit{CLWK} reflect the difference.
We devise two loss functions to weakly supervise them. The latent weight scores are expected to be a continuous value thus we consider this task as a regression task rather than a classification task. Mean Squared Error (MSE) is adopted as the loss function. 
To obtain the pseudo ground truth, we use the F1 score to measure the closeness between a context utterance and the response on the word level.
As for its values, for \textit{CLWR}, the context utterance with the maximum F1 score is tagged as 1 and the rest of the utterances to be 0. It is worth noting that the last utterance, i.e., the post, is always 1 because it has been proven crucial for response generation \cite{sankar2019neural}. 

For training \textit{CLWK}, we use the same method for constructing the pseudo ground truth. The only difference is that \textit{CLWK} is built for the knowledge latent weight, so we produce the most relevant knowledge for the F1 score calculation. First of all, we use the TF-IDF approach to retrieve from the knowledge sentences by taking the response as the query, i.e., ranking the knowledge sentence by \emph{TF-IDF}(knowledge sentence, response).\footnote{\scriptsize TF-IDF($\cdot$) is the TF-IDF (term frequency-inverse document frequency) function. IDF is obtained by the individual dataset.} The top-1 ranked sentence based on the TF-IDF is treated as the most important knowledge sentence, being tagged as \textit{Top1-RK}. Secondly, similar to \textit{CLWR}, the context utterance with the maximum F1(Context Utterance, \textit{Top1-RK}) is used to supervise \textit{CLWK}. 
Formally,
%
\begin{flalign}
\small
    \scalemath{1.0}{\textit{GT}_{\textit{CLWR(i)}} = 
    \begin{cases}
     & \text{1, } \text{if c}_i \text{= argmax(F1(c}_i \text{, R))} \\
     &  \text{1, }  \text{if c}_i \text{= post } \\
     & \text{0, } \text{otherwise} \\
    \end{cases}}
\end{flalign}
%
\begin{flalign}
\small
    \scalemath{1.0}{\textit{GT}_{\textit{CLWK(i)}} = 
    \begin{cases}
     & \text{1, } \text{if c}_i \text{= argmax(F1(c}_i \textit{, Top1-RK))} \\
     & \text{1, } \text{if c}_i \text{= post} \\
     & \text{0, } \text{otherwise} 
    \end{cases}}
\end{flalign}
in which $c_i$ means each context utterance. Then, we define the loss function to be:
\begin{equation} \label{loss:cwpr}
\small
    \scalemath{1.0}{\textit{Loss}_{\textit{CLWR}} = \textit{MSE}(\textit{CLWR}, \textit{GT}_{\textit{CLWR}})}
\end{equation}
\begin{equation} \label{loss:cwpk}
\small
    \scalemath{1.0}{\textit{Loss}_{\textit{CLWK}} = \textit{MSE}(\textit{CLWK}, \textit{GT}_{\textit{CLWK}})}
\end{equation}
where $\textit{GT}_{\textit{CLWR}}$ and $\textit{GT}_{\textit{CLWK}}$ are the pseudo ground-truth context utterance scores for response generation and knowledge selection tasks respectively.

\subsection{Knowledge Latent Weight Generator} \label{sec:kwp_generator}
The knowledge Latent Weight generator is designed to generate a knowledge latent weight (KLW). It begins with a knowledge latent vector, which is a word embedding indexed by a fixed index of 1. Note that the knowledge latent word embedding is different from the context latent embedding. 

\vspace{0.05in}
\noindent {\bf Latent Weight Enhanced Attention.}
Latent Weight Enhanced Attention (LWE Attention) is built on top of the standard attention by considering the latent weights. Originally, the attention is calculated between two sequence representations from the word level (shown in Eq.~\ref{eq:attention}). The LWE Attention takes sentence-level scores, i.e., the latent weights, into consideration. By this, the Eq.~\ref{eq:attention} is then changed to be:
\begin{equation} \label{eq:pe_attention}
\small
    \scalemath{1.0}{\textit{LWE} \ \textit{Attention(Q, K, V) = LW} \times \textit{softmax} \left ( \frac{\textit{QK}^\textit{T}}{\sqrt{\textit{d}}} \right ) \textit{V}}
\end{equation}
where $\textit{LW}$ stands for latent weights. $\textit{LW}$ will be different when predicting responses and generating knowledge latent weight. Namely, in the \textit{KLW} generator, the $\textit{LW}$ is replaced with \textit{CLWK}. In the Decoder, it is changed to \textit{CLWR} and \textit{KLW}, which will be introduced in Sec.~\ref{sec:decoder}.

\vspace{0.05in}
\noindent {\bf Context \& Knowledge Dependency} (CK-Dep for short).
Prior studies \cite{focus_prabhumoye2021focused, three_stage-liu2021three} consider the context and knowledge dependency by stacking a context cross-attention and a knowledge cross-attention from word level. We also leverage the stacked architecture and consider the context sentence-level weights (through LWE Attention), i.e., the context LWE cross-attention module and the knowledge cross-attention module in \textit{KLW} generator.

\vspace{0.05in}
\noindent {\bf Weak Supervision on \textit{KLW}.}
After going through the CK-Dep operation, the Latent Vector is processed by the Latent Weight Head module to get the \textit{KLW}. The knowledge generally contains richer information than the context \cite{ted_zheng2019enhancing, skt-Kim-2020-ICLR}. For context we take the top-1 ranked utterance as the pseudo ground truth $\textit{GT}_{\textit{CLWK}}$. However, for knowledge, we set a hyper-parameter $N$ to get the pseudo ground truth knowledge sentences $\textit{GT}_{\textit{KLW}}$. Namely, the top $N$ ranked knowledge sentences are considered to be the ground truth for supervising \textit{KLW}.
\begin{equation}
\small
    \scalemath{1.0}{\textit{GT}_{\textit{KLW(i)}} = 
    \begin{cases}
     & \text{1, } \text{k}_i \text{=Top N argmax(TF-IDF(k, R))} \\
     & \text{0, } \text{otherwise} 
    \end{cases}}
\end{equation}
where, $k_i$ is each knowledge sentence. With different Extrema and
Greedy scores of the ZRKGC model are lower than
the CKL. This means although the generated re-
sponses of the ZRKGC model are closer to the
ground truth response on average, it can not seman-
tically capture the most important words.
Third, in terms of the diversity scores, the pro-
posed CKL does not improve over other models but
we expect that the method can be improved by re-
fining the use of latent weights which are currently
normalized between 0 and 1 and multiplied by the
word attention scores. Despite of CKL model’s gen-
erated responses being not the most diverse among
all compared models, our human evaluation results
reveal that CKL is preferred by the 5 annotators
with moderate agreement, in terms of relevance,
coherence, informativeness, and overall preference
(Appendix B.2).
3https://github.com/ellenmellon/DIALKI
4https://github.com/neukg/KAT-TSLF
5https://github.com/shrimai/
Focused-Attention-Improves-Document-Grounded-GenerationN for the individual dataset, the performance could vary. We discuss the effect of top-N in Appendix~\ref{sec:top_n_knowledge}. Similar to the \textit{CLWR} and \textit{CLWK}, we also consider the \textit{KLW} generation as a regression task, and the loss function is:
\begin{equation} \label{loss:kwp}
\small
    \scalemath{1.0}{\textit{Loss}_{\textit{KLW}} = \textit{MSE(KLW,} \textit{GT}_{\textit{KLW}})}
\end{equation}

\vspace{-0.1in}
\subsection{Decoder and Training} \label{sec:decoder}
The Decoder is a BART decoder but equipped with LWE Attention. In the `Context \& Knowledge LWE Cross-Attention' module in Figure~\ref{fig:architecture}, the context and knowledge representations are multiplied by the corresponding latent weights. Namely, the $\textit{LW}$ in Eq.~\ref{eq:pe_attention} will be replaced by \textit{CLWR} and \textit{KLW} when dealing with context and knowledge in the Decoder. Formally, the Eq.~\ref{eq:pe_attention} is instantiated to be:
\begin{equation} \label{eq:decoder_pe}
\small
    \scalemath{1.0}{\begin{aligned}
      \textit{PE} \ & \textit{Attention(Q, K, V)} = \\
    & \sum_{i=1}^m{\textit{CLWR}_i \times \textit{softmax}\left ( \frac{\textit{QK}_i^\textit{T}}{\sqrt{\textit{}d}} \right ) \textit{V}_i} + \\
      & \sum_{j=1}^l{\textit{KLW}_j \times \textit{softmax}\left ( \frac{\textit{QK}_j^\textit{T}}{\sqrt{\textit{d}}} \right ) \textit{V}_j}
    \end{aligned}}
\end{equation}
where, $K_i$ and $V_i$ means $i$-th context utterance. $K_j$ and $V_j$ denote $j$-th knowledge sentence. $\textit{CLWR}_i \in \mathcal{R}^{1 \time 1}$ and $\textit{KLW}_j \in \mathcal{R}^{1 \time 1}$ stand for the corresponding context and knowledge latent weights.
The loss function for response generation is a Negative Log Likelihood loss (NLL).
%
\begin{equation} \label{loss:nll}
\small
    \scalemath{1.0}{\textit{Loss}_{\textit{NLL}} = -\sum_{i=1}^{\textit{L}} \textit{log} \; \textit{p(R}_\textit{t}|\textit{R}_{<t}\textit{, C, K}). }
\end{equation}
in which, $L$ is the maximum length of the response, $t$ is the $t$-th token to be generated and $R_{<t}$ denotes the generation steps prior to $t$.

\vspace{0.05in}
\noindent {\bf Aggregation of Loss Functions.}
In this paper, we have four different loss functions, including Eq.~\ref{loss:cwpr}, Eq.~\ref{loss:cwpk}, Eq.~\ref{loss:kwp} and Eq.~\ref{loss:nll}. Previous studies simply aggregate different loss functions by either an addition operation \cite{itdd_li2019incremental, ktwm-zheng2021knowledge} or setting hyper-parameters to do a weighted sum \cite{dialki-wu2021dialki}, which are sub-optimal. 
\citet{awl1-kendall2018multi} propose a principled approach to multi-task learning which weighs multiple loss functions by considering the homoscedastic uncertainty of each task. This Automatic Weighted Loss (AWL) allows the model to simultaneously learn various quantities with different units or scales in various settings. Adopting this strategy, we define our final loss as follows:
\begin{equation} \label{loss:awl}
\small
\scalemath{1.0}{\begin{aligned}
  \textit{Loss} &= \frac{1}{2\delta^2_1}\textit{Loss}_{\textit{CLWR}} + \frac{1}{2\delta^2_2}\textit{Loss}_{\textit{CLWK}} + \\
  & \frac{1}{2\delta^2_3}\textit{Loss}_{\textit{KLW}} + \frac{1}{2\delta^2_4}\textit{Loss}_{\textit{NLL}} + \textit{log(}\delta_1\delta_2\delta_3\delta_4\textit{)}
\end{aligned}}
\end{equation}
The final goal is to minimize the objective with respect
to $\delta_1, \delta_2, \delta_3$ and $\delta_4$ as learning the relative weight of the four different losses.

\section{Experiment}

\subsection{Datasets, Settings, and Metrics }

\noindent {\bf Datasets.} Following previous research practices \citep{doha-prabhumoye2021focused, kat-liu2021three, itdd_li2019incremental, zhao2020knowledge, three_stage-liu2021three}, we use two public datasets: Wizard of Wikipedia (WoW for short. \citet{wizard_dinan2018wizard}) and CMU-DoG \citep{cmudog-zhou2018dataset} to conduct our experiments. 
We introduce them in detail in Appendix~\ref{appendix:datasets}.

\vspace{0.05in}
\noindent \textbf{Experimental Settings.} All of the experiments are conducted with the same hyper-parameter settings as described in Appendix~\ref{appendix:exp_setting}.

\vspace{0.05in}
\noindent {\bf Metrics.} 
As used in previous works \cite{firstknowledge_Ghazvininejad, drd-Zhao2020Low-Resource, itdd_li2019incremental, ktwm-zheng2021knowledge}, we employ BLEU \cite{bleu-papineni_etal_2002_bleu}, Rouge \cite{rouge_lin_2004_rouge}, Diversity \cite{diversity_li2015diversity} and embedding-based metric, BOW Embedding \citep{embedding_liu2016not} as the metrics. Among them, BLEU calculates N-grams co-occurrence between two sequences. Rouge measures the number of overlapping units such as word sequences, and word pairs between the generated sequence and the ground truth sequence. Diversity score counts distinct N-grams number divided by the total number of the generated corpus. BOW Embedding metric leverages the pre-trained word vector to calculate the similarity between sequences from the semantic space. Meanwhile, \citet{bleu2isbetter-liu2016not, meteor-banerjee2005meteor} suggest that compared with the other metrics, BLEU2, and embedding-based metrics have a better correlation with human assessment, and thus in this paper, we take BLEU2 and embedding-based measurements as the main metrics for discussion. 

To have a better understanding of the proposed model, we also conducted a human evaluation (reported in Appendix~\ref{appendix:human_evaluation}).

\subsection{Baseline Approaches} \label{sec:baselines}

We compare our CKL model with six baselines.
\vspace{0.1in}

\noindent {\bf ITDD} \cite{itdd_li2019incremental} proposes an incremental Transformer architecture to improve context coherence and knowledge correctness. 

\noindent {\bf DRD} \cite{drd-Zhao2020Low-Resource} proposes a disentangled response decoder to isolate parameters that depend on knowledge-grounded dialogues from the entire generation model. 

\noindent {\bf ZRKGC} \cite{zrkgc-li2020zero} treats the knowledge as latent variables so that the model can estimate the knowledge representation distribution from the latent space.\footnote{\scriptsize \url{https://github.com/nlpxucan/ZRKGC}} 

\noindent {\bf DIALKI} \cite{dialki-wu2021dialki} proposes a knowledge identification model to provide dialogue-contextualized passage encodings and locate knowledge that is relevant to the conversation.\footnote{\scriptsize \url{https://github.com/ellenmellon/DIALKI}} 
\noindent {\bf KAT} \cite{kat-liu2021three} devises a three-stage architecture to get better context inner-relationship, knowledge representation, and interaction between context and knowledge.\footnote{\scriptsize \url{https://github.com/neukg/KAT-TSLF}} 

\noindent {\bf DoHA} \cite{doha-prabhumoye2021focused} focuses on building a context-driven representation of the document and enabling specific attention to the information in the document.\footnote{\scriptsize \url{https://github.com/shrimai/Focused-Attention-Improves-Document-Grounded-Generation}}

\begin{table*}[htb]
   \scriptsize
  \centering
  
  \caption{Automatic evaluation results on Wizard of Wikipedia (WoW) test seen and CMU-DoG datasets. * means significant test value with \textit{p < 0.05}, compared to the CKL. Note that the results of ITDD and DRD are copied from the papers, so they do not have significant test results. `w/o' means without a certain module for the ablation study while `red.' means reduced to a subset of the whole dataset for training. All values are expressed as percentages (\%).}
  \label{tab:automatic}
  \begin{tabular}{l|llllllllll}
    \toprule
    \multirow{2}*{Models}  & \multicolumn{10}{c}{Wizard of Wikipedia test seen}  \\

    ~ & BLEU-1 & BLEU-2 & BLEU-3 & BLEU-4  & Rouge-L & Div-1 & Div-2 & Average & Extrema  & Greedy \\
    \midrule
    ITDD \cite{itdd_li2019incremental}       & 15.80  &  7.10 & 4.00  & 2.50  &  -  & -  & - & - & - & - \\
    DRD \cite{drd-Zhao2020Low-Resource}    & 21.80  &  11.50  & 7.50 & 5.50  & -  & -   & -  &  - & -  &  -  \\
    ZRKGC \cite{zrkgc-li2020zero}    & 23.80*  &  8.80*  &  4.04*  & 1.95*  & 16.86*  & 5.44*  &  22.66*  & {\bf 72.32*}  &  40.40*  &  41.67*    \\
    KAT \cite{kat-liu2021three}   &  16.92* &  9.28*  & 6.04* & 4.37*   & 16.41*  & {\bf 14.66*}  & {\bf 45.99}  & 67.86*  & 39.06*  & 39.37*     \\
    DoHA \cite{doha-prabhumoye2021focused}    &  23.19* &  11.70*  &  6.99* & 4.65*   & 21.32*  & 7.44  & 31.47  & 69.91*  & 40.91*  & 41.64*           \\
    DIALKI \cite{dialki-wu2021dialki}  &  25.00*   &  13.72*  &  9.09* & 6.68*  &   22.10*  & 9.33  & 41.71  &  70.43*  &  42.31*  &  41.73*       \\
    \hline
    CKL     &  {\bf 27.29} &  {\bf 15.80}  & {\bf 10.99}  &  {\bf 8.41}  & {\bf 23.96}  & 9.03  &  36.36  & 71.11  & 42.95  &  42.54    \\
    w/o $\textit{Loss}_{\textit{KLW}}$     &  26.01 & 14.80  & 10.16  & 7.68   & 23.25  & 9.37  & 37.27  & 70.69* & 42.56  & 42.02*     \\
    w/o $\textit{Loss}_{\textit{CLWR}}$      & 26.57  & 15.43  & 10.76  & 8.22    & 23.75  & 9.31  & 37.38 & 71.05  & {\bf 43.50*} &  42.39  \\
    w/o $\textit{Loss}_{\textit{CLWK}}$      & 26.67  & 15.49  & 10.75  & 8.21    & 23.81  & 9.37  & 37.11*  & 70.99 & 43.37*  & 42.50     \\
    w/o \textit{CK-Dep}      & 26.68  & 15.51  & 10.76  & 8.21    & 23.82  & 9.25  & 36.99*  & 71.18 & 43.35*  & {\bf 42.66}     \\
    \hline
    red. \textit{1/2 training data}      & 25.01* & 13.81*  & 9.23*  &  6.85*   & 22.37*   & 9.04  &  35.79* &  70.94 & 42.30*  &  42.41    \\
    red. \textit{1/4 training data}     & 23.48* &  12.48*  &  8.03* & 5.77*    &  21.17*  &  8.56* &  34.45* &  70.76 & 42.16*  &  42.63    \\
    \hline
    \hline
    \multirow{2}*{Models}  & \multicolumn{10}{c}{CMU-DoG test set}  \\
    ~ & BLEU-1 & BLEU-2 & BLEU-3 & BLEU-4  & Rouge-L & Div-1 & Div-2 & Average & Extrema  & Greedy \\
    \midrule
    ITDD     &  9.50 & 3.60  & 1.70  & 0.90  & -  & -  & -  &  -& -  & -  \\
    DRD      &  15.00  & 5.70  &  2.50 & 1.20  & -   & -  & -  &  -& -  & -  \\
    ZRKGC    & 17.35*  & 5.68*  &  2.31* & 1.00*  &  13.05*  &  1.34* &  8.26*	 & {\bf 66.26*}  &  31.42* &  37.91*  \\
    KAT       & 13.53*  & 5.81*  &  2.81* &  1.49*   & 11.98*  &  {\bf 4.19*} & 17.60*  & 63.72* & 35.18*  & 37.72*  \\
    DoHA     & 17.02*  & 6.95*  &  3.30* & 1.72*   & 14.41*  & 2.39*  & 12.14*  & 65.69 &  35.51* &  39.26*  \\
    DIALKI     & 15.83*  &  6.41* & 3.10   & 1.69  & 14.64*  &  3.43* & {\bf 20.43*}  & 63.57* & 34.89*  & 37.60* \\
    \hline
    CKL     & {\bf 17.74}  & {\bf 7.91}  &  4.11 & 2.29   &  15.87 & 2.30  & 11.10  & 65.63 & 35.81  & 39.46  \\
    w/o $\textit{Loss}_{\textit{KLW}}$     &  17.05 & 7.66  &  4.06 &  2.34   & 15.75  & 2.47  & 11.69  & 65.57* & 35.74  &  {\bf 39.48}    \\
    w/o $\textit{Loss}_{\textit{CLWR}}$      & 17.02  &  7.62 &  3.98 & 2.23  &   {\bf 15.93}  &  2.34 &   11.32 & 65.19 &  35.71* &  39.20*    \\
    w/o $\textit{Loss}_{\textit{CLWK}}$      & 17.33  & 7.80  & {\bf 4.14}  & {\bf 2.38}  &   15.91  & 2.37*  &  11.30 & 65.65 & {\bf 35.83}  &  39.47    \\
    w/o \textit{CK-Dep}      &  17.16 &  7.64 & 4.01   & 2.27  & 15.86  & 2.37*  & 11.67*  & 65.30* & 35.63  &  39.10*    \\ \hline
    red. \textit{1/2 training data}     & 16.99* &  7.25* & 3.67*   & 2.02*  &  14.84*  &  2.32* & 11.51*   & 65.07*  & 34.68*  &  38.30*    \\
    red. \textit{1/4 training data}      & 17.22* & 7.27*  & 3.61*   &  1.94*  &  14.20*  &  2.07* &  10.35*  & 65.11*  &  34.90* &   39.09*   \\
  \bottomrule
\end{tabular}
\end{table*}

\subsection{Experiment Results}
\label{exp:results}

\noindent \textbf{Main Results.}
The experimental results are shown in Table~\ref{tab:automatic}. Because of the page limitation, we report the results of the WoW test unseen set (with similar trends) in Appendix~\ref{appendix:wow_test_unseen}.
First, based on BLEU and Rouge-L scores, the proposed CKL models perform consistently better than the baseline approaches. This reflects that the results from the CKL share more consecutive tokens with the ground truth responses. Looking closely at the BLEU-2 scores, the CKL's results are improved by large margins compared to the best results of the baseline approaches (DIALKI); they are around 15\% better for the WoW test seen (improving from 13.72\% to 15.80\%) and around 14\% better for the CMU-DoG dataset. 

Second, for the embedding-based metrics, the CKL is better than most of the baseline models except for the ZRKGC model on the Embedding Average measurement.
However, the Extrema and Greedy scores of the ZRKGC model are lower than the CKL. This means although the generated responses of the ZRKGC model are closer to the ground truth response on average, it can not semantically capture the most important words. 

Third, in terms of the diversity scores, the proposed CKL does not improve over other models but we expect that the method can be improved by refining the use of latent weights which are currently normalized between 0 and 1 and multiplied by the word attention scores.
Despite of CKL model's generated responses being not the most diverse among all compared models, our human evaluation results reveal that CKL is preferred by the 5 annotators with moderate agreement, in terms of relevance, coherence, informativeness, and overall preference (Appendix~\ref{appendix:human_evaluation}).

\vspace{0.05in}
\noindent \textbf{Ablation Study.}
Four downgraded versions of CKL are provided. (1) w/o $\textit{Loss}_{\textit{KLW}}$, i.e., removing the knowledge latent weight loss function; (2) w/o $\textit{Loss}_{\textit{CLWR}}$ (deleting the context latent weight supervision for response generation); (3) w/o $\textit{Loss}_{\textit{CLWK}}$ (deleting the context latent weight supervision for knowledge prediction); and (4) w/o \textit{CK-Dep} removes the context-knowledge dependency when generating \textit{KLW}, i.e., removing the Context LWE Cross-Attention module from the \textit{KLW} Generator in Figure~\ref{fig:architecture}. From Table~\ref{tab:automatic} we can see that all of the BLEU-2 scores decrease. For the WoW dataset, when removing $\textit{Loss}_{\textit{KLW}}$ and $\textit{Loss}_{\textit{CLWK}}$, the BLEU-2 gets the lowest score among all of the ablation experiments, indicating the importance of knowledge latent weights generation. In terms of the CMU-DoG dataset, which has patterns different from the WoW test seen, w/o $\textit{Loss}_{\textit{CLWR}}$ decreases the most. Thus, correctly identifying context seems more crucial than knowledge selection for the CMU-DoG dataset. We presume that  CMU-DoG's knowledge sentences are complementary, i.e., different knowledge sentences contain similar information, resulting in the context recognition showing more importance. We further verify this assumption in Appendix~\ref{sec:top_n_knowledge}. Other metrics decreased to varying degrees but most remain better than the baseline approaches. On the whole, the full version of the CKL performs the best.

\vspace{0.05in}
\noindent \textbf{Low-Resource Experiments.}
In order to test the CKL's robustness, we also conduct experiments on low-resources scenarios. 
From Table~\ref{tab:automatic}, the BLEU-2 scores of the CKL model with half of the training data are respectively 13.81\% and 7.25\% on WoW test seen and CMU-DoG datasets, outperforming the best baseline models (DIALKI with 13.72\% on WoW, and DoHA with 6.95\% on CMU-DoG). 
Appendix~\ref{appendix:low_resource} further shows although as the scale of the training set decreases, the performance drops gradually, our proposed CKL model still performs reasonably well with limited training data.

\begin{figure}[t]
  \centering
    \includegraphics[width=0.9\linewidth]{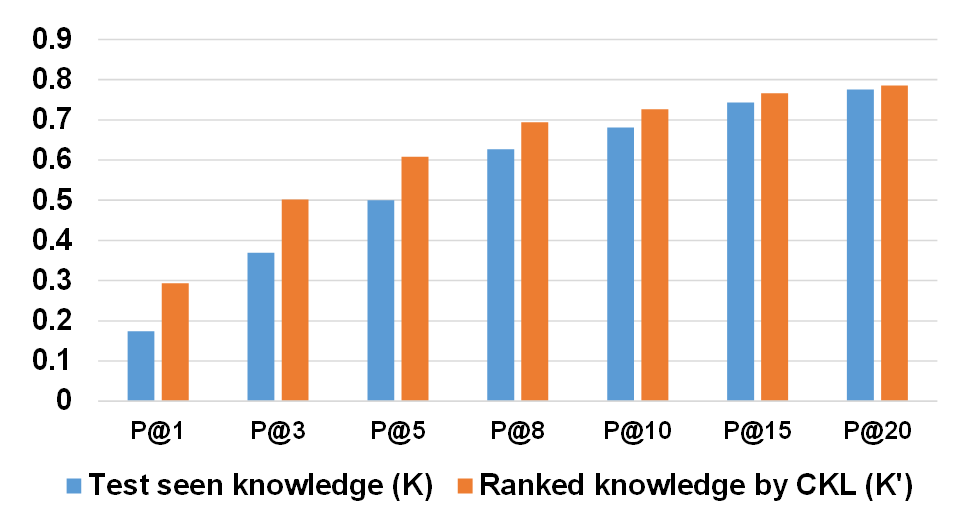}
  \caption{P@N scores for WoW test seen original knowledge set and the re-ranked knowledge set by CKL.}
  \label{fig:patn}
  \vspace{-0.1in}
\end{figure}

\vspace{0.05in}
\noindent \textbf{Latent Weight Analysis.}
To illustrate the effectiveness of the proposed CKL model, we demonstrate (1) knowledge re-ranking by the knowledge latent weights and (2) Spearman's Correlation between the knowledge latent weights and the pseudo ground truth scores. We use the WoW test seen set for illustration. The same patterns are found for WoW test unseen and CMU-DoG datasets.

WoW and CMU-DoG datasets provide a set of initial knowledge, designated by $K$. The predicted knowledge latent weights by CKL are scores for each knowledge sentence that can be used to rank knowledge and obtain re-ranked knowledge set $K'$. We construct pseudo ground truth knowledge order by using the response as the query to retrieve from the knowledge named $K_{\textit{GT}}$. At this point, the top 1 ranked knowledge sentence in $K_{\textit{GT}}$ is the most relevant to the response,  \textit{Top1Klg}.
We use \textit{P@N} as the metric to evaluate the precision. For a sample, we calculate the percentage of \textit{Top1Klg} included within the top N-ranked knowledge sentences.

Figure~\ref{fig:patn} shows the results: for the original knowledge order \textit{K}, \textit{P@1} is about 17.5\%; for \textit{K}'   \textit{P@1} score is around 30\%. For each N, the \textit{P@N} for \textit{K}' is higher than for \textit{K}. That confirms that the latent weight modules can improve the relevance scoring of knowledge sentences. In Appendix~\ref{appendix:case_study} we also provide a qualitative case study of the rankings.

To further analyze the effectiveness of the latent weights \textit{CLWR}, \textit{CLWK} and \textit{KLW}, we calculate Spearman's Correlation between each weight group and the corresponding pseudo ground truths which have been elaborated on in \checkthis{Sec.~\ref{sec:method}} (e.g., for \textit{CLWR} we calculate the Spearman's Correlation between \textit{CLWR} and $\textit{GT}_{\textit{CLWR}}$).
We use the ZRKGC model to provide weights for knowledge sentences without context weight (ZRKGC does not provide it) and obtain Spearman's Correlation with the pseudo-ground truth $\textit{GT}_{\textit{KLW}}$. We compare the resulting Spearman's Correlation coefficients with those of the CKL and CKL's ablated models. The results are shown in Table~\ref{tab:spearman}. For \textit{KLW}, the coefficients are higher than for ZRKGC by a large margin. For CKL's ablated models, the coefficients are lower than for the full CKL model. For instance,  \textit{KLW} correlation score 0.0966 for `w/o $\textit{Loss}_{\textit{KLW}}$' is much lower than the score 0.37 for CKL. This further demonstrates all supervised modules are helpful to the entire model.

\begin{table}[t]
   \scriptsize
  \center
  
  \caption{Spearman's Correlation between latent weights and the pseudo ground truths. * means t-test value with $p < 0.05$, in comparison with the proposed \checkthis{CKL}. }
  \label{tab:spearman}
\begin{tabular}{l|lll}
    \toprule
     Models  & KLW & CLWR & CLWK  \\
    \midrule
    ZRKGC       &  0.1001*  & -  & - \\
    CKL    & 0.3700  & 0.6697 &  0.6455   \\
    w/o $\textit{Loss}_{\textit{KLW}}$   &  0.0966*  &  0.6699  & 0.6455   \\
    w/o $\textit{Loss}_{\textit{CLWR}}$   &  0.3585  & 0.4658*   & 0.6455  \\
    w/o $\textit{Loss}_{\textit{CLWK}}$       & 0.3694  & 0.6696  & 0.5769* \\
    w/o \textit{CK-Dep}        & 0.3732 & 0.6696 &  0.4816*\\
    
  \bottomrule
\end{tabular}
\vspace{-0.05in}
\end{table}

\section{Conclusion}

Past studies on capturing context and knowledge relationships to boost the quality of response generation models were restricted to coarse-grain characterization through context-knowledge cross-attention. 
%
{In this paper, we describe the Contextual Knowledge Learning (CKL) method for response generation. We propose our CKL model by using two latent vectors which are trained to capture the relationship between context, responses, and the `best knowledge' (identified through a pre-defined default retrieval process by taking the response as the query) as well as the relationship between contextual knowledge and responses. The trained latent vectors are used to generate latent weights that enhance the traditional attention operation by multiplying them with the token-level attention scores. Furthermore, we leverage the weak supervision technique to jointly train the latent weights production as well as the response generation.
With these two mechanisms, the CKL has the flexibility of influencing the learning process and has demonstrated superior performance against six strong baselines. 
Future work will explore more diverse use of latent vectors and latent weights as part of the learning process.   }

\section{Limitations}
The proposed CKL can automatically produce the scores for each utterance in the context and each sentence in the knowledge. However, it is constrained by the total length of the input sequence. CKL takes BART as its foundation, thus the bottleneck of BART limits the upper ability of CKL. BART requests the input length to be 1,024, which means the CKL can receive at most 1,024 tokens at a time. For some samples, the concatenation of the context and knowledge contains far more than 1,024 but is truncated to fit with the length requirement. In those cases, the CKL cannot get enough information, resulting in the sub-optimal performance of the CKL.

\section{Ethical Statement}
We are aware of how personal identities, expectations, norms, and values may influence our study. Our datasets were released by previous studies and are publicly available sources. Since researchers had no interactions with human subjects in this case, our ethics review board did not consider this part of our research as human subjects research. The only part of the study that involved human subjects was a crowd-sourcing human evaluation study (Appendix~\ref{appendix:human_evaluation}), where participants were paid at least the minimum wage. All of our analysis were based on aggregated data without tracking down to individuals.

\section*{Acknowledgments}
We would like to thank Benjamin Towle for his generous assistance in providing valuable suggestions and proofreading support. This work is partly supported by Engineering and Physical Sciences Research Council (EPSRC Grant No. EP/S515528/1, 2102871). The Titan V used for this research was donated by the NVIDIA Corporation. All content represents the opinion of the authors, which is not necessarily shared or endorsed by their respective employers and/or sponsors.

\bibliography{anthology,custom}
\bibliographystyle{acl_natbib}

\appendix

\section{Experimental Setup}
To enable the reproduction of results, we make our code publicly available at \url{https://github.com/tonywenuon/acl2023-ckl}.

\subsection{Datasets} \label{appendix:datasets}
\vspace{0.1in}
\noindent {\bf Wizard of Wikipedia Dataset}. 
The dataset \citep{wizard_dinan2018wizard} includes knowledge annotated by workers from the Amazon Mechanical Turk (AMT) platform. Each worker was given a set of background knowledge retrieved based on the dialogue history from Wikipedia articles to assess. The dataset is split into train/validation/test sets. For the validation and test sets, based on the topic existence in the train set, there are two versions: seen set (the topics exist in the train set) and the unseen set (some new topics are not contained in the train set). The original dataset can be obtained at http://parl.ai. The train/seen validation/seen test sets' sizes are 74,092/3,939/3,865. For the seen validation/seen test sets, the sizes are 3,927/3,924.

\vspace{0.1in}
\noindent {\bf CMU-DoG Dataset}.
This dataset is proposed by \citet{cmudog-zhou2018dataset}. It also employs workers from the AMT and its conversations are mainly about movies. They are required to exchange ideas about movies. The original dataset can be obtained from the published paper\footnote{\scriptsize \url{https://github.com/festvox/datasets-CMU\_DoG}}. \citet{itdd_li2019incremental} also released a CMU-DoG dataset\footnote{\scriptsize \url{https://github.com/lizekang/ITDD}}, which has tokenization to all of the source and target sequences. In our paper, we use the ITDD version. The train/validation/test sets consist of 66,332/3,269/10,502 samples.

\subsection{Experiment Settings} \label{appendix:exp_setting}
In our experiment, the BART-base model\footnote{\scriptsize \url{https://huggingface.co/facebook/bart-base}} is used. The maximum source length is 1024 tokens and 64 tokens for the target length. For the number of context utterances, we use the latest 10 utterances. In terms of knowledge, we use the maximum source length rather than the number of knowledge sentences to determine how many to incorporate. The learning rate is set to be 5e-5. All of the experiments are trained for 10 epochs on a single TITAN V GPU. The proposed CKL model needs about 20 hours for training on the Wizard of Wikipedia dataset and about 8 hours on the CMU-DoG dataset.

\section{Complementary Experimental Results}

\begin{table*}[htb]
   \scriptsize
  \centering
  
  \caption{Automatic evaluation results on Wizard of Wikipedia test unseen set. * means significant test value with \textit{p < 0.05}, compared to the CKL. `w/o' means without a certain module for the ablation study. All values are expressed as percentages (\%).}
  \label{tab:wow_unseen}
    \vspace{-0.05in}
  \begin{tabular}{l|llllllllll}
    \toprule
    \multirow{2}*{Models}  & \multicolumn{10}{c}{Wizard of Wikipedia test unseen}  \\
    ~ & BLEU-1 & BLEU-2 & BLEU-3 & BLEU-4  & Rouge-L & Div-1 & Div-2 & Average & Extrema  & Greedy \\
    \midrule
    ITDD \cite{itdd_li2019incremental}     &  13.40  & 4.70  &  2.10 & 1.10     & - & - & - & - & - & - \\
    DRD \cite{drd-Zhao2020Low-Resource}   &  20.70 & 10.10  & 6.20 & 4.30   & - & - & - & - & - & - \\
    ZRKGC \cite{zrkgc-li2020zero}  & 23.30*   &  8.50*  &  3.94* & 1.94*    & 16.81*  &  3.48* &  15.78*  &  {\bf 71.93*}  & 39.11*  & 41.43*  \\
    KAT \cite{kat-liu2021three}    &  16.49*    &  8.49* &  5.30* & 3.67*  & 15.60*  & {\bf 9.84*}  & {\bf 31.13*}  & 67.02*  & 37.20*  & 38.55*  \\
    DoHA \cite{doha-prabhumoye2021focused}  & 22.28*   &  10.53*  & 6.16* & 4.10*  &  20.04*  & 5.04*  & 21.97*  &  68.61*  &  37.96*  &  40.23*     \\
    DIALKI \cite{dialki-wu2021dialki}   & 25.26*    & 13.96*  & 9.36* & 6.96*  &  22.02*  &  5.28  &  25.67 & 69.67*  &  40.94* &  41.39* \\
    \hline
    CKL    & {\bf 27.68}  &  {\bf 16.05} & {\bf 11.22} &  {\bf 8.62}   & {\bf 23.93}  &  4.89  & 18.60 &  70.36  & {\bf 41.93}  &  {\bf 41.86} \\
    w/o $\textit{Loss}_{\textit{KLW}}$     &  26.57 &  15.13 & 10.38  & 7.86  &   23.39 & 5.31  & 20.25  &  69.99* & 41.51 &     41.62   \\
    w/o $\textit{Loss}_{\textit{CLWR}}$      &  26.87 & 15.48  & 10.77  & 8.21	 & 23.53  & 5.04  & 19.26  & 69.97* & 41.68  &  41.73    \\
    w/o $\textit{Loss}_{\textit{CLWK}}$      & 26.49  & 15.13  & 10.47  & 7.97    & 23.33  & 5.16  & 19.67  & 69.85* & 41.59  &  41.56*    \\
    w/o \textit{CK-Dep}      & 27.01 & 15.55  &  10.85 &  8.34 &   23.67 & 5.20  &  19.93 & 70.19 & 41.74  &   41.83   \\
  \bottomrule
\end{tabular}
\end{table*}

\subsection{Results on WoW Test Unseen Set} \label{appendix:wow_test_unseen}
From Table~\ref{tab:wow_unseen}, it is clear that the CKL performs best among all of the experiments in terms of the BLEU and Rouge scores. Like the results of the Wizard of Wikipedia (WoW) test seen set, the diversity scores and the Embedding-Average score are slightly worse than KAT and ZRKGC respectively. Remarkably, the BLEU-2 score improves from 13.96\% to 16.05\%, which is even higher than that of the WoW test seen set, which indicates the effectiveness of the CKL model.

\begin{table*}
    \centering
  \scriptsize
  \caption{Human evaluation on Wizard of Wikipedia test seen and CMU-DoG datasets. The values except for Kappa are in percentage (\%).}
  \label{tab:human}
  \begin{tabular}{l|cccc|cccc}
    \toprule
    \multirow{2}*{CKL vs. DIALKI} & \multicolumn{4}{c|}{Relevance} & \multicolumn{4}{c}{Coherence}      \\

    ~  & Win & Loss & Tie & Kappa  & Win & Loss & Tie & Kappa  \\
     \midrule
     WoW test seen & 41.34  &  25.25 &  33.41  &  0.30     &  42.66  &  24.26 &  33.08  & 0.33    \\
     CMU-DoG & 49.51 & 20.79 & 29.70 & 0.39       & 44.55 & 26.73 & 28.72 & 0.45  \\
     \midrule
    \multirow{2}*{CKL vs. DIALKI}  & \multicolumn{4}{c|}{Informativeness} & \multicolumn{4}{c}{Overall Preference}    \\
     ~  & Win & Loss & Tie & Kappa  & Win & Loss & Tie & Kappa  \\
     \midrule
     WoW test seen & 43.23  & 24.26 &  32.51  &  0.31      & 36.30     &  22.28  &  41.42  &  0.39  \\
     CMU-DoG     & 48.51 & 39.61 & 11.88 & 0.42       & 50.50 & 38.61 & 10.89 & 0.41     \\
  \bottomrule
\end{tabular}
\end{table*}

\subsection{Human Evaluation}  \label{appendix:human_evaluation}
We also conducted human evaluation by deploying users through the crowd-sourcing Amazon MTurk platform. 5 AMT workers were employed to assess samples from 4 perspectives, i.e., Relevance, Coherence, Informativeness, and Overall Preference. Following \citet{ccta-ling2021context}, the four criteria are referred to as: {\bf Relevance} - whether the generated response is relevant to the given context. {\bf Coherence} - whether the generated response is a coherent and meaningful continuation of the dialogue. {\bf Informativeness} - how many new and diverse expressions do the generated responses introduce. {\bf Overall Preference} - personal preference between two responses. 

To do the human evaluation, we randomly select 100 samples from the outputs of the proposed CKL and DIALKI (the best-performing baseline model) for the Wizard of Wikipedia and CMU-DoG datasets respectively. Then we use the Amazon Mechanical Turk platform for assessment. The assessors are asked to select a response that they preferred from two models on different perspectives (relevance, coherence informativeness, and overall), and they are also allowed to consider both responses are equal to the given context, i.e., `Tie' in Table~\ref{tab:human}. 

The results are shown in Table~\ref{tab:human}, from which we can see that for all of the four criteria, the proposed CKL is better than the DIALKI. This indicates that the CKL model improves in terms of relevance, coherence, and informativeness. We calculate Fleiss' Kappa \cite{fleiss1973equivalence} for each criteria. The resulting Kappa scores are around 0.4, which indicates a moderate agreement among the assessors. From Table~\ref{tab:human}, we can observe that our proposed CKL model outperforms the best baseline DIALKI model from all perspectives.

\begin{table*}
  \scriptsize
  \center
  \caption{Wizard of Wikipedia test seen \& unseen and CMU-DoG evaluation results on low-resource scenarios. * means significant test value with $p < 0.05$, in comparison with the full version of \checkthis{CKL}. All values are expressed as percentages (\%).}
  \label{tab:low_resource}
  \begin{tabular}{l|llllllllll}
    \toprule
    \multirow{2}*{Models}  & \multicolumn{10}{c}{Wizard of Wikipedia test seen}  \\

    ~ & BLEU-1 & BLEU-2 & BLEU-3 & BLEU-4  & Rouge-L & Div-1 & Div-2 & Average & Extrema  & Greedy \\
    \midrule
    Full training data   & 27.29 &  15.80 &  10.99 &  8.41   & 23.96   & 9.03  & 36.36  & 71.11  & 42.95  & 42.54     \\
    1/2 training data      & 25.01* & 13.81*  & 9.23*  &  6.85*   & 22.37*   & 9.04  &  35.79* &  70.94 & 42.30*  &  42.41    \\
    1/4 training data     & 23.48* &  12.48*  &  8.03* & 5.77*    &  21.17*  &  8.56* &  34.45* &  70.76 & 42.16*  &  42.63    \\
    1/8 training data      & 19.43* & 10.38*  &  6.86* & 5.10*    &  18.32*  & 9.58*  & 36.35*  &  67.84*  &  40.88* & 41.08*     \\
    1/16 training data       &  20.02*& 9.77*  &  6.04* & 4.26*    &  18.84*  &  8.33* & 31.82*   & 67.12*  &  40.03* & 40.05*      \\
    Zero training data     & 9.01* & 4.14*  & 2.34*  &  1.54*   &  11.44*  & 5.66*  &  25.63* & 56.87*  & 34.90*  &  35.45*    \\
    
    \hline
    \hline
    
     \multirow{2}*{Models}  & \multicolumn{10}{c}{Wizard of Wikipedia test unseen}  \\
    ~ & BLEU-1 & BLEU-2 & BLEU-3 & BLEU-4  & Rouge-L & Div-1 & Div-2 & Average & Extrema  & Greedy \\
    \midrule
    Full training data    & 27.68 & 16.05  & 11.22   & 8.62    &  23.93  &  4.89 & 18.60   & 70.36  &  41.93 & 41.86     \\
    1/2 training data     &  25.41* & 14.09*  &  9.46*  &  7.06*  &  22.41*	  &5.14   & 20.17   &  69.88* & 40.48*	  &   41.75   \\
    1/4 training data      & 23.25* &  12.25* &  7.94*  &  5.76*  &  20.87*  &  5.31* &  20.93*  &  69.69* & 40.55*  & 41.70     \\
    1/8 training data     & 18.81*  & 10.09*  & 6.79*   &  5.11*  &  18.04*  & 5.48*  &  20.58*  & 65.88*  &  38.82* & 39.92*     \\
    1/16 training data      & 19.96*  & 9.64*  & 6.04*   & 4.33*   &  18.52*  & 4.49*  & 17.04*   &  65.56*	 & 38.12*   & 39.00*      \\
    Zero training data     & 8.76* & 3.96*  &  2.24*  &  1.47*  &   11.32* & 3.55*  &  17.10*  & 56.36*  & 33.54*  &  35.60*    \\

    \hline
    \hline

    \multirow{2}*{Models}  & \multicolumn{10}{c}{CMU-DoG}  \\

    ~ & BLEU-1 & BLEU-2 & BLEU-3 & BLEU-4  & Rouge-L & Div-1 & Div-2 & Average & Extrema  & Greedy \\
    \midrule
    Full training data   & 17.74 &  7.91 &  4.11  &  2.29  &  15.87  & 2.30  &  11.10	  &  	65.63  & 35.81  &    39.46  \\
    1/2 training data     & 16.99* &  7.25* & 3.67*   & 2.02*  &  14.84*  &  2.32* & 11.51*   & 65.07*  & 34.68*  &  38.30*    \\
    1/4 training data      & 17.22* & 7.27*  & 3.61*   &  1.94*  &  14.20*  &  2.07* &  10.35*  & 65.11*  &  34.90* &   39.09*   \\
    1/8 training data     & 15.47* &  6.11*	 &   2.94* & 1.59*   &  13.74*  & 2.59*  &  13.00*  & 62.11*  & 33.53*	 &  37.16*    \\
    1/16 training data      & 15.01* & 5.68*  &  2.61*  & 1.32*	   &  13.97*  & 2.43*	  & 12.22*    &  62.05* & 34.06*  &   36.72*   \\
    Zero training data     & 7.21* &  2.51* &  0.96*  &  0.42*  &  8.47* & 2.92*  & 18.99*   &  62.65* &  30.26* & 35.61*     \\

  \bottomrule
\end{tabular}
\end{table*}

\subsection{Low-Resource Complementary Experiments} \label{appendix:low_resource}

Table~\ref{tab:low_resource} shows the results of the complementary experiments for low-resource training. We can clearly see the effectiveness of the proposed CKL. As the scale of the training set decreases, the performance drops gradually. 
However, when the scale of the training data goes down to less than 1/4 of the original training data, the performance decreases more dramatically. Our proposed CKL model performs reasonably well with limited training data, but models with sufficient amount of training data are still preferred.

\begin{figure}[t]
  \centering
    \includegraphics[width=\linewidth]{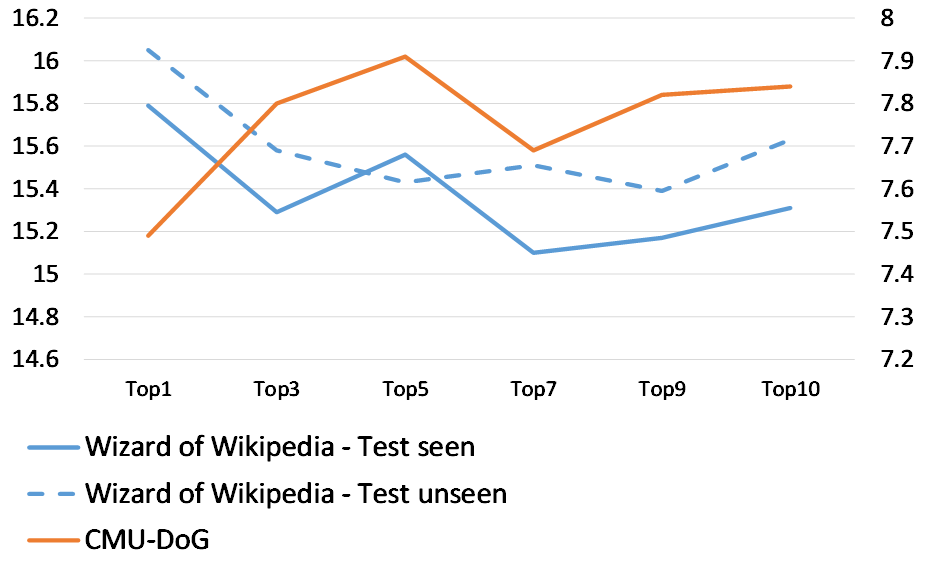}
  \caption{BLEU-2 scores (Y-axis) for WoW and CMU-DoG datasets with different numbers of top N retrieved knowledge sentences being the ground truth.}
  \label{fig:top_n_knowledge}
\end{figure}

\subsection{Effect of Top-N Retrieved Knowledge} \label{sec:top_n_knowledge}

In Sec.~\ref{sec:kwp_generator}, we set top \textit{N} retrieved knowledge sentences as the ground truth for obtaining $\textit{GT}_{\textit{KLW}}$. We will discuss how different \textit{N} affects the performance. As can be seen in Figure~\ref{fig:top_n_knowledge}, we investigate from the top 1 to top 10 retrieved knowledge sentences. It is clear that for the WoW dataset, using the first retrieved knowledge gets the best results, while for the CMU-DoG dataset, the top-5 group peaks. That indicates that in the WoW dataset, the knowledge other than the top 1 retrieved sentence contains limited useful information. However, the knowledge in the CMU-DoG dataset complements each other. This also explains why when removing $\textit{Loss}_{\textit{CLWR}}$, the BLEU-2 decreases most for CMU-DoG (see ablation study in Sec~\ref{exp:results}).

\begin{table*}[t]
\centering
  \footnotesize
  \begin{tabular}[H]{l|l}
    \toprule
    \multirow{13}*{\bf Good Case}  & \multicolumn{1}{c}{\bf Context, Target and Generated Responses}  \\
    \cline{2-2}
    ~& ~ \\
    ~ & {\bf Context: } Pop music. I sure am a fan of pop music these days. It is so much fun to listen to!   \\
    ~ & {\bf Ground Truth Target: }  Pop music is a genre of popular music that originated in its modern form in the United \\
    ~ & States, and yes it is fun to listen to.  \\
    ~& ~ \\
    \cline{2-2}
    ~& ~ \\
    ~ & {\bf ZRKGC: } I'm a fan of pop music, but it is not the most popular genre. \\
    ~ & {\bf KAT: } I love pop music too! it originated in the US and UK in the 1950s.  \\
    ~ & {\bf DoHA: } Pop music is a genre of music that originated in the US and Canada.  \\
    ~ & {\bf DIALKI: } Pop and rock are popular music. I like both but I prefer pop music.  \\
    ~ & {\bf CKL: } Pop music is a genre of popular music that originated in its modern form in the United States and the \\
    ~ & UK during the mid-1950s.  \\
    ~& ~ \\
    \hline
    \hline
    \multirow{18}*{\bf Bad Case}  & \multicolumn{1}{c}{\bf Context, Target and Generated Responses}  \\
    \cline{2-2}
    ~& ~ \\
    ~ & {\bf Context: }    \\
    ~ & {\bf A: } Swimming is consistently among the top public recreational activities, and in some countries, swimming  \\
    ~ & lessons are a compulsory part of the educational curriculum.   \\
    ~ & {\bf B: } Swimming has been consistently the most popular recreational activity.    \\
    ~ & {\bf A: } It is a fantastic exercise that people can do at almost any age.   \\
    ~ & {\bf Ground Truth Target: } Especially since it mainly relies on the body's natural buoyancy.  \\
    ~& ~ \\
    \cline{2-2}
    ~& ~ \\
    ~ & {\bf ZRKGC: } Swimming is consistently the most popular recreational activity. I don't know about it, but in some  \\
    ~ & countries, they have to be a time. \\
    ~ & {\bf KAT: } It is! Humans can hold their breath underwater and undertake rudimentary locomotive swimming within \\ 
    ~ & weeks of birth.  \\
    ~ & {\bf DoHA: } It is! And in some countries, swimming lessons are a compulsory part of the educational curriculum. \\
    ~ & {\bf DIALKI: } It is amazing that people can do it almost every day, whether it be for recreation, sport, exercise, or \\
    ~ & survival. \\
    ~ & {\bf CKL: } I think it is the most popular recreational activity.  \\
    ~& ~ \\
    
  \bottomrule
\end{tabular}
  \caption{A good case and a bad case of the proposed CKL model. In the base case, `A' and `B' denote the two Amazon MTurk workers exchanging their opinions.}
  \label{tab:case_study}
\end{table*}

\section{Case Study} \label{appendix:case_study}
In order to qualitatively demonstrate the results generated by the CKL and the baseline models, we report a good case and a bad case which are generated by the CKL model. The good case is selected by this criterion for a sample: if a knowledge weight is the highest among all of the knowledge sentences and it is also the ground truth knowledge (see latent weight analysis in Sec.\ref{exp:results}), it is viewed as good. On the contrary, if the knowledge sentence is the ground truth knowledge but predicted to have the lowest latent score, it is a bad case. From Table~\ref{tab:case_study}, we can see that in the good case, if the knowledge is predicted correctly, the CKL-generated response is very close to the ground truth target. The other responses produced by the baseline approaches tend to be generic. In terms of the bad case, due to the lack of important information used in the target, the CKL response does not share keywords with the ground truth target even though it is a proper answer for the given context. This comes to the one-to-many problem which describes that many responses are reasonable for a certain dialogue scenario by the natural language. This problem is out of this paper's scope.


\end{document}